\documentclass[journal]{IEEEtran}
\usepackage{cite}
\ifCLASSINFOpdf
\else
\fi

\usepackage[numbers, compress]{natbib}
\usepackage[utf8]{inputenc} % allow utf-8 input
\usepackage[T1]{fontenc}    % use 8-bit T1 fonts
\usepackage[colorlinks,linkcolor=blue,urlcolor=blue,citecolor=blue]{hyperref}% hyperlinks
\usepackage{url}            % simple URL typesetting
\usepackage{booktabs}       % professional-quality tables
\usepackage{amsfonts}       % blackboard math symbols
\usepackage{nicefrac}       % compact symbols for 1/2, etc.
\usepackage{microtype}      % microtypography

\usepackage{xspace}
\usepackage{color}
\usepackage{stmaryrd}
\usepackage{amsthm}
\usepackage[pdftex]{graphicx}
\graphicspath{{paper_figures/}}
\usepackage{multirow}

\usepackage{algorithm}
\usepackage{algorithmic}
\usepackage{arydshln}
\usepackage{threeparttable}

\newcommand{\eg}{\textit{e.g.}}
\usepackage{amsmath,amsfonts,bm}

\def\eqref#1{equation~\ref{#1}}
\def\1{\bm{1}}

\DeclareMathAlphabet{\mathsfit}{\encodingdefault}{\sfdefault}{m}{sl}
\SetMathAlphabet{\mathsfit}{bold}{\encodingdefault}{\sfdefault}{bx}{n}

\newcommand{\xhdr}[1]{{\noindent\bfseries #1}.}

\newcommand{\cut}[1]{}

\begin{document}
\title{EchoEA: Echo Information between Entities and Relations for Entity Alignment}

\author{
  Xueyuan Lin, Haihong E\thanks{Corresponding Author}, Wenyu Song, Haoran Luo \\
  Department of Computer Science, \\
  Beijing University of Posts and Telecommunications, Beijing, China\\
  \texttt{linxy59@mail2.sysu.edu.cn, ehaihong@bupt.edu.cn} \\
  \texttt{swy9834@bupt.edu.cn, luohaoran@bupt.edu.cn} \\
}

\ifCLASSOPTIONpeerreview
\markboth{IEEE Transactions on Audio, Speech and Language Processing}%
{EchoEA: Echo Information between Entities and Relations for Entity Alignment}
\else
\markboth{IEEE Transactions on Audio, Speech and Language Processing}%
{Xueyuan \MakeLowercase{\textit{et al.}}: EchoEA: Echo Information between Entities and Relations for Entity Alignment}
\fi



\maketitle

\begin{abstract}
Entity alignment (EA) plays an important role in automatically integrating knowledge graphs (KGs) from multiple sources.
Recent approaches based on Graph Neural Network (GNN) obtain entity representation from relation information and have achieved promising results. Besides, more and more methods introduce semi-supervision to ask for more labeled training data.
However, two challenges still exist in GNN-based EA methods: (1) Deeper GNN Encoder: The GNN encoder of current methods has limited depth (usually 2-layers). (2) Low-quality Bootstrapping: The generated semi-supervised data is of low quality.
In this paper, we propose a novel framework, Echo Entity Alignment (EchoEA), which leverages 4-levels self-attention mechanism to spread entity information to relations and echo back to entities.
Furthermore, we propose attribute-combined bi-directional global-filtered strategy (ABGS) to improve bootstrapping, reduce false samples and generate high-quality training data.
The experimental results on three real-world cross-lingual datasets are stable at around 96\% at hits@1 on average, showing that our approach not only significantly outperforms the state-of-the-art GNN-based methods, but also is universal and transferable for existing EA methods.
\end{abstract}


% \fi
\IEEEpeerreviewmaketitle

\section{Introduction}
% \IEEEPARstart{K}{nowledge}
Knowledge graphs (KGs) consist of nodes (entities) and edges (relationships between entities, attributes of entities), which have been widely applied for knowledge-driven tasks such as question answering, recommendation system, and search engine.
The key step in integrating multi-source KGs is to infer their intersection, that is, to align equivalent entities.
Therefore, entity alignment (EA) task has attracted increasingly attention in recent years.

The popular EA framework includes three steps. (1) Embedding the entity into a low-dimensional vector space using knowledge graph embedding (KGE) models. (2) Calculating the similarity matrix of the entities to be aligned. (3) Obtaining the predicted target entities based on the similarity matrix.

Variety of papers introduce KGE methods to EA task to help capture the semantics hidden in KGs.
Traditional KGE methods include Translation-based models~\cite{TransE, TransR, MTransE}, rotating models~\cite{RotatE}, polar coordinate models~\cite{HAKE}, bilinear models~\cite{RESCAL,DistMult,ComplEx}, hyperbolic geometry models~\cite{Poincare,MuRP}, Convolution-based models~\cite{ConvE}, Capsule-based models~\cite{CapsE} and so on. These models have already been proved to be effective in link prediction tasks.
However, \citet{TransEdge} noticed that most of those models designed for link prediction perform even worse than TransE~\cite{TransE} in EA task.
\citet{RREA} explained that it is because transformation matrix which transforms entity embeddings into relation specific ones is difficult to comply with the orthogonal property. That is to say, transformation matrix should be orthogonal. To obstacle the problem, they design a Graph Neural Networks (GNN) based network to obtain entity embedding. Therefore, in this paper, we will also propose our GNN-based model and keep orthogonal transformation matrix.

Popular GNN-based methods include GCN~\cite{GCN}, GraphSAGE~\cite{GraphSAGE}, GAT~\cite{GAT}, DGI~\cite{DGI} and GIN~\cite{GIN}. For EA task, previous approaches pay more attention to the connectivity of the graph, ignoring the relation types between the head and tail entities, the direction of the relations, the contribution of entity information to the relation, etc.
GCN-Align~\cite{GCN-Align} only focuses on whether there are edges between entities, without considering the number of relation types.
RREA~\cite{RREA} regards relation as reflection action and intergrates reflection into Graph Attention Network (GAT).
Recently, RAGA~\cite{RAGA} considers this, but does not further distinguish the contribution of different parts of the relations to the entities.

Comparing GNN-based with KGE-based methods, GNN-based methods require a large amount of pre-aligned seeds.
It is common to generate more seeds for training with bootstrapping or iterative strategies~\cite{BootEA, MRAEA}.
But these methods extremely depend on the network performance. In this paper, we propose a more efficient bootstrapping strategy.

However, recent studies show that there are two critical challenges still existing as follow:

\paragraph{Deeper GNN Encoder}

For EA task, most GNN-base methods take 2-layers GCN or 2-layers another GNN as encoder.
The state of the art papers~\cite{RREA,EMGCN,RAGA} design multi-layers GCN to encode KG topology at multiple orders.
But it is interesting that 2-layers GCN performs the best in their experiments.
On the one hand, from the research in the field of GNN, it is already a consensus that multi-layers GCN is trapped in oversmoothing and overfitting. Previous work introduce Highway Network or skip-connections to make the network deeper, but the depth is still limited.
On the other hand, to our best knowledge, a deeper GNN network means not only broader vision to digest KG topology but also high level of noise to degrade the quality of the embeddings.
It is of significance to build a deeper and better-performance GNN encoder.
In this paper, we design the entity GNN encoder of four levels attention, which successfully reaches five layers.

\paragraph{Low-quality Bootstrapping}

The bootstrapping methods hold the view that the model should perform better if it becomes more confident about its predicted results, which are fed back to the model as training data.
BootEA~\cite{BootEA} and MRAEA~\cite{MRAEA} are classic methods to adopt iterative or bootstrapping strategies to build semi-supervised models.
However, since the quality of generated training data strongly depends on GNN models, the data contains amounts of false samples.
Previous methods ignored all the negative samples and only appended the positive ones to their training set.
On the contrary, this paper makes use of local alignment and global alignment to enhance the quality of generated positive and negative samples, which is more efficient than before.

% Local alignment (LA) means that each source entity selects the target entity from the candidate set with replacement, so there is a disadvantage that multiple source entities point to the same target entity. Global alignment (GA)~\cite{RAGA, CEA} means that each source entity selects the target entity from the candidate set without replacement. There is only one-to-one alignment, more stringent. Prior work did not use these two methods simultaneously, and the framework proposed in this paper includes both.

\label{section:introduction-EA}

We summarise our contributions as follows:

\begin{itemize}

  \item \textbf{Model}: We design a novel GNN-based model, \emph{Echo}, to further encode the entity feature. The model is built with four levels attention networks, which has addressed several key issues from prior methods.
  \item \textbf{Strategy}: We propose an iterative strategy, \emph{Attribute-combined Bi-directional Global-filtered Strategy (ABGS)}, to generate semi-supervised data of high-quality. We utilise negative samples for bootstrapping, not just positive samples.
  \item \textbf{Learning}: Our iterative strategy \emph{ABGS} reduces false samples. It brings about 50\% decrease in false positive rate and false negative rate when compared with the state-of-the-art iterative method MRAEA~\cite{MRAEA}.
  \item \textbf{Experiments}: We conduct extensive experiments on three most common public cross-lingual datasets to demonstrate the efficacy of our model. Our proposed iterative model significantly outperforms all the state-of-the-art (GNN-based and iterative) approaches and reaches about 96\% on average on \emph{Hits@1}.
  \item \textbf{Open Source}: The source code is available at \url{https://github.com/LinXueyuanStdio/EchoEA}.
\end{itemize}

\section{Related Work}\label{sec:relatedwork}
\xhdr{GNN-Based Methods}
The famous one to apply GNNs to entity alignment task is GCN-Align~\cite{GCN-Align}, which introduces multi-layers (actually 2-layers) vanilla GCN encoder and contrastive loss function.
However, since vanilla GCN is too simple to model heterogeneous relations in KGs, GCN-Align is unable to utilise more information about relations.
\citet{RDGCN} proposed RDGCN to construct a dual relation graph which regards relation as node and entity as edge.
In basic KGs the information spreads from relations to entities, while in dual relation graph the information spreads from entities to relations.
\citet{MRAEA} noticed the importance of relation types. They assigned weight coefficients according to relation types to distinguish entity features better.
To fulfill both relational differentiation and dimensional isometry criteria at the same time, RREA~\cite{RREA} incorporated relational reflection transformation in GNNs.
To utilise multiple relations for more reasonable entity representation sufficiently, \citet{RAGA} proposed RAGA with relation-aware graph attention networks to implicitly model the interaction between entities and relations.
From GCN-Align to RAGA, there is an obvious trend that the interaction of relations and entities is becoming increasingly important while the depth of GNN encoder is limited to 2.
% Our framework EchoEA dynamically represents relations and entities, and merges them to final entity representations, which is further interaction modeling.

\xhdr{Iterative Training}
GNN-based methods is \emph{supervised}, which requires plenty of pre-aligned seeds.
However, in practice, the aligned seeds are often inadequate due to the high cost of manual annotations and the huge size of KG.
To expand training data, some recent studies~\cite{MRAEA, BootEA} adopt iterative or bootstrapping strategies to build semi-supervised models.
BootEA~\cite{BootEA} employed an alignment editing method to reduce error accumulation during iterations.
To simplify things, MRAEA~\cite{MRAEA} considered the asymmetric nature of cross-lingual alignment direction, thus reduces the error propagation problem brought by adding falsely aligned pairs into the training set of next epoch.
Similarly, RREA~\cite{RREA} generated extra positive samples via semi-supervision which improves the performance by an average of 6\% on \emph{Hits@1}.
However, the generated data is of low quality and highly depends on the model performance.
Besides, these methods only generate positive samples and drop negative samples.
In this paper, we show that negative sampling in iterative strategy is vital. Additionally, we enhance our strategy with attribute information and alignment methods to generate high quality data.

\xhdr{Attribute Information}
Attribute information is an important part of KGs.
JAPE~\cite{JAPE} was the first to utilise attribute information for EA task but gaining low performance due to high levels of inconsistency and linguistic differences.
GCN-Align~\cite{GCN-Align} considered only the types of attributes and ignores their values.
After that, EMGCN~\cite{EMGCN} leveraged the advance in machine translation to reconcile the attribute information of cross-lingual KGs without the need for any human-related supervision data.
Inspired by these work, we inject attribute information to the iterative strategy.

\xhdr{Alignment Methods}
To generate alignment, each source entity selects the target entity from the candidate set.
If the selection is with replacement, the alignment is named Local Alignment (LA).
Otherwise, the alignment is named Global Alignment (GA)~\cite{RAGA, CEA} because it is one-to-one alignment.
CEA~\cite{CEA} adopts deferred acceptance algorithm (DAA) to guarantee stable matches, which is an approximate GA method and can significantly reduce time complexity.
RAGA~\cite{RAGA} calculates a fine-grained similarity matrix by summing the weights of each entity aligned in two directions (left-to-right and right-to-left).
Since GA predicts more accurate results, it will contribute to iterative strategy and reduce false samples.

\section{Preliminaries}
A knowledge graph can be represented as $KG=\{E, R, A, V, T^R, T^A\}$, where $E$ is the set of entities, $R$ is the set of relations, $A$ is the set of attributes, $V$ is the set of attribute values, $T^R=\{ (h,r,t) \in E \times R \times E \}$ is the set of relation triples, and $T^A=\{ (e,a,v) \in E \times A \times V \}$ is the set of attribute triples. Relation triple $(h,r,t)$ consists of head entity $h \in E$, relation $r \in R$ and tail entity $t \in E$. Attribute triple $(e,a,v)$ consists of entity $e \in E$, attribute $a \in A$ and attribute value $v \in V$.

Given $KG_1=\{E_1, R_1, A_1, V_1, T^R_1, T^A_1\}$, $KG_2=\{E_2, R_2, A_2, V_2, T^R_2, T^A_2\}$ and pre-aligned seeds $P_{\text{seed}}=\{(e_{1}, e_{2}) \in E^{\text{seed}}_{1} \times E^{\text{seed}}_{2}, e_1 \leftrightarrow e_2\}$ where $\leftrightarrow$ represents equivalence, $|P_{\text{seed}}| = |E^{\text{seed}}_1| = |E^{\text{seed}}_2|$ and $E^{\text{seed}}_i \subseteq E_i (i=1,2)$, we split pre-aligned seeds to training set $P_{\text{train}}=\{(e_{1}, e_{2}) \in E^{\text{train}}_{1} \times E^{\text{train}}_{2}, e_1 \leftrightarrow e_2\}$ and testing set $P_{\text{test}}=\{(e_{1}, e_{2}) \in E^{\text{test}}_{1} \times E^{\text{test}}_{2}, e_1 \leftrightarrow e_2\}$ where $P_{\text{train}} \cap P_{\text{test}} = \emptyset$,$P_{\text{train}} \cup P_{\text{test}} = P_{\text{seed}}$. EA task is to predict $? \in E^{\text{test}}_{2}$ for $(e_{1}, ?)$ given $e_1 \in E^{\text{test}}_{1}$ (left-to-right prediction) and $? \in E^{\text{test}}_{1}$ for $(?,e_{2})$ given $e_2 \in E^{\text{test}}_{2}$(right-to-left prediction). A bootstrapping strategy is to generate likely entity pairs $P'=\{(e_{1}, e_{2}) \in E^{\text{test}}_{1} \times E^{\text{test}}_{2}\}$ to expand training set $P_{\text{train}}$. For $(e_1,e_2)$ in $P'$, equivalence may not hold between $e_1$ and $e_2$.

\section{EchoEA Framework}\label{section:Echo}
\begin{figure*}
  \centering
  \vspace{-3mm}
  \includegraphics[width=0.8\textwidth]{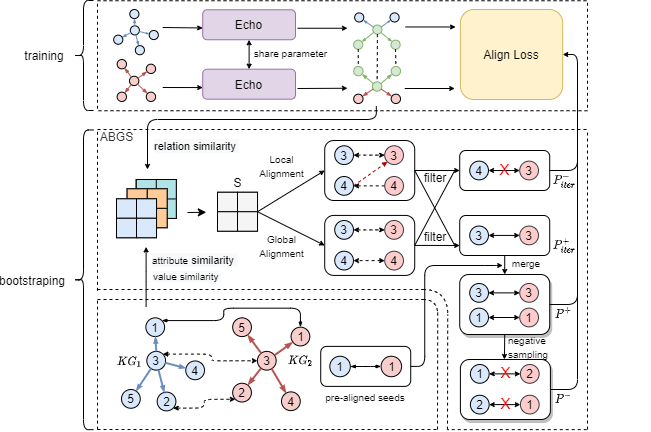}
  \caption{Overall architecture of EchoEA. The framework containes tow parts, training and bootstrapping. In training, we encode entity feature for each KG via \emph{Echo} by sharing parameters and then apply Align Loss to train the model. At the same time, the entity feature is copied to calculate relation similarity for bootstrapping. In bootstrapping, firstly, we combine relation similarity, attribute similarity and value similarity to one similarity matrix to prepare for alignment. Then, with Local Alignment to bi-directionally align the entites and Global Alignment to filter, we merge these likely pairs with pre-aligned seeds and perform negative sampling to generate high quality entity pairs. Finally, these generated pairs are fed back for training. When the training process stops, the results of Global Alignment is the final results.}
  \vspace{-3mm}
  \label{Fig-EchoEA}
\end{figure*}

In this work, We propose EchoEA framework which takes GNN-based \emph{Echo} to encode entity representation and Attribute-combined Bi-directional Global-filtered Strategy (\emph{ABGS}) to generate extra data. Figure~\ref{Fig-EchoEA} depicts the overall architecture of EchoEA.

\subsection{Echo Model}

\begin{figure*}
  \centering
  % \vspace{-3mm}
  \includegraphics[width=0.9\textwidth]{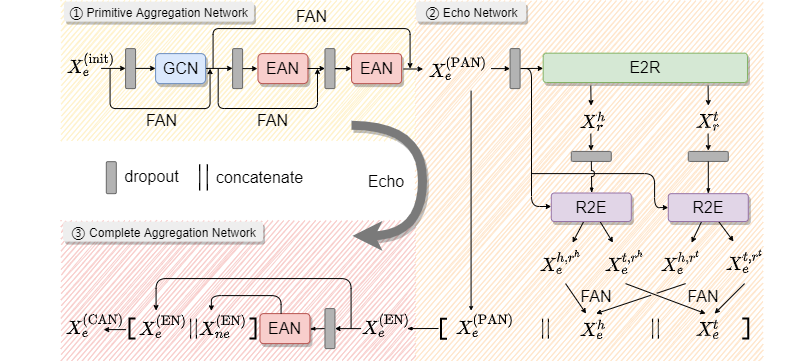}
  \caption{Overall architecture of \emph{Echo} model. It is built with Graph Convolution Network (GCN), Entity-level Attention Network (EAN), Feature-level Attention Network (FAN), Entity-to-Relation Attention Network (E2R), Relation-to-Entity Attention Network (R2E), Primitive Aggregation Network (PAN), Echo Network (EN), Complete Aggregation Network(CAN). $\mathbf{X}_e^{(\text{init})} \in \mathbb{R}^{|E|\times d_e}$ is the initial entity feature matrix, $|E|$ is entity count of KG and $d_e$ is entity embedding dimension. $\mathbf{X}_e^{(\text{PAN})}$ is the output entity feature matrix of PAN. $\mathbf{X}_r^{h},\mathbf{X}_r^{t} \in \mathbb{R}^{|R|\times d_r}$ is the output relation feature matrix of E2R, where $|R|$ is relation count of KG and $d_r$ is relation embedding dimension. With input $\mathbf{X}_r^{h}$ and $\mathbf{X}_e^{(\text{PAN})}$, R2E computes $\mathbf{X}_e^{h,r^h}, \mathbf{X}_e^{t,r^h}$. Similarly, with input $\mathbf{X}_r^{t}$ and $\mathbf{X}_e^{(\text{PAN})}$, the other R2E computes $\mathbf{X}_e^{h,r^t}, \mathbf{X}_e^{t,r^t}$. Then one FAN combines the head views and the other FAN combines the tail veiws. $\mathbf{X}_e^{(\text{EN})}$ is the output entity feature matrix of EN. $\mathbf{X}_{ne}^{(\text{EN})}=\text{EAN}(\mathbf{X}_e^{(\text{EN})})$. $\mathbf{X}_e^{(\text{CAN})}$ is the output entity feature matrix of CAN.}
  \vspace{-3mm}
  \label{Fig-EchoModel}
\end{figure*}

To obtain entity representation, we propose \emph{Echo} including 3 parts: Primitive Aggregation Network (PAN), Echo Network (EN) and Complete Aggregation Network (CAN):

\begin{equation}
  \mathbf{X}_e^{(\text{Echo})}=\text{Echo}(\mathbf{X}_e^{(\text{init})}) = \text{CAN}(\text{EN}(\text{PAN}(\mathbf{X}_e^{(\text{init})})))
\end{equation}

where $\mathbf{X}_e^{(\text{init})} \in \mathbb{R}^{|E|\times d_e}$ is the initial entity feature matrix, $|E|$ is entity count of KG and $d_e$ is entity embedding dimension, $\mathbf{X}_e^{(\text{Echo})} \in \mathbb{R}^{|E|\times d_e'}$ is the encoded entity embedding matrix, $d_e'$ is output dimension, $\text{Echo}(.)=\text{CAN}(\text{EN}(\text{PAN}(.)))$ is \emph{Echo} model which encodes entity feature through a sequence of networks PAN, EN and CAN. Figure~\ref{Fig-EchoModel} depicts the network structure of \emph{Echo}.

%? 分3个level：feature level attention, entity level attention, relation level attention

\xhdr{Graph Convolution Network (GCN)} GCN is harnessed to generate basic structural representations of entities.

\begin{equation}
  \mathbf{X}_e^{(\text{GCN})}=\sigma(\hat{\mathbf{D}}^{-\frac{1}{2}}\hat{\mathbf{M}}\hat{\mathbf{D}}^{-\frac{1}{2}}\mathbf{X}_e^{(\text{init})}\mathbf{W})
\end{equation}

where $\mathbf{X}_e^{(\text{GCN})}$ is the output of GCN, $\sigma(.)$ is an activation function, $\hat{\mathbf{M}}=\mathbf{M}+\mathbf{I}$ is an adjacency matrix with self-connections, $\hat{\mathbf{D}}$ is degree matrix, $\mathbf{W}$ is $d^{(in)}\times d^{(out)}$ weight matrix. Inspired by RREA~\cite{RREA}, we keep $d^{(in)}=d^{(out)}=d_e$ to avoid destroying the shape.

GNN-based EA methods \cite{GCN-Align} usually use 2-layer GCN to encode entity features. Some papers \cite{EMGCN,RAGA} claim that they use multi-layers GCN instead, and 2-layers GCN performs best in experiments. We agree that GCN can utilise structure information in KG. However, GCN suffers from oversmoothing and overfitting when deeper. To overcome the problem, we introduce attention mechanism of four levels (entity-level, feature-level, entity-to-relation and relation-to-entity) to further encode entity feature.

\xhdr{Entity-level Attention Network (EAN)} To capture the importance of neighbor entities, we employ Entity-level Attention Network to assign attention score to each neighbor entity. To simplify computation and resist overfitting, only a vector parameter is trainable.

\begin{equation}
  \alpha_{ij}=\frac{\exp(\text{LeakyReLU} (\mathbf{a}^T[\mathbf{x}_i || \mathbf{x}_j]))}{\sum_{j' \in N_i}\exp(\text{LeakyReLU} (\mathbf{a}^T[\mathbf{x}_i || \mathbf{x}_{j'}]))}
\end{equation}
\begin{equation}
  \mathbf{x}_i^{(l)}=\sum_{j'\in N_i}\alpha_{ij}\mathbf{x}_i^{(l-1)}
\end{equation}

where $\mathbf{x}_i^{(l)}$ is the vector embedding of entity $e_i$ in $l$-th layer, $N_i$ is the id set of neighbor entities of $e_i$, $\alpha_{ij}$ represents the attention weight from $e_j$ to $e_i$, $\text{LeakyReLU}(.)$ is activation function, $||$ represents the concatenate operation, $\mathbf{a}$ is one-dimensional trainable vector to map the $2d_e$-dimensional input into a scalar.

\xhdr{Feature-level Attention Network (FAN)} To fuse different features, the state of the art directly combines feature-specific embeddings via addition, multiplication, concatenate operation or Highway Network. These operations either ignore or manually assign the weights of features, which can be inapplicable when the number of features increases, or the importance of certain features varies greatly under different settings. Therefore, we propose Feature-level Attention Network to adaptively determine the weight of each feature.

\begin{equation}
  \alpha = \text{Sigmoid}(\mathbf{a}^T[\mathbf{x}^{(a)} || \mathbf{x}^{(b)}])
\end{equation}
\begin{equation}
  \mathbf{x}^{(out)}=\alpha \mathbf{x}^{(a)}+(1-\alpha) \mathbf{x}^{(b)}
\end{equation}

where $\mathbf{x}^{(out)}$ is the output entity embedding vector, $\mathbf{x}^{(a)}$($\mathbf{x}^{(b)}$) is the entity embedding vector of fature $a$($b$), $\text{Sigmoid}(.)$ is activation function, $\mathbf{a}$ is another one-dimensional trainable vector to map the $2d_e$-dimensional input into a scalar.

\xhdr{Entity-to-Relation Attention Network (E2R)} To overcome overfitting, we give up asigning embedding vector for each relation because the embedding parameters may also lead to overfitting. Instead, we propose Entity-to-Relation Attention Network to dynamically compute relation representation as follow:

\begin{equation}
  \alpha_{ijk}=\frac{\exp(\text{LeakyReLU}(\mathbf{a}^T[\mathbf{x}_i^h\| \mathbf{x}_j^t]))}{\sum_{e_{i'}\in \mathcal{H}_{\mathbf{r}_k}}\sum_{e_{j'}\in\mathcal{T}_{e_{i'}\mathbf{r}_k} }\exp(\text{LeakyReLU}(\mathbf{a}^T[\mathbf{x}_{i'}^h\| \mathbf{x}_{j'}^t]))}
\end{equation}
\begin{equation}
  \mathbf{r}_k^{h}=\sum_{e_{i}\in \mathcal{H}_{\mathbf{r}_k}}\sum_{e_{j}\in\mathcal{T}_{e_{i}\mathbf{r}_k} }\alpha_{ijk} \mathbf{x}_i^h
\end{equation}

where $\mathcal{H}_{\mathbf{r}_k}$ is the set of head entities of relation $\mathbf{r}_k$, $\mathcal{T}_{e_{i}\mathbf{r}_k}$ is the set of tail entities of head entity $e_i$ and relation $\mathbf{r}_k$, $\alpha_{ijk}$  represents attention weight from head entity $e_i$ to relation $\mathbf{r}_k$ based on head entity $e_i$ and tail entity $e_j$, $\mathbf{a}$ is another one-dimensional trainable vector to map the $2d_r$-dimensional input into a scalar. Compared with asigning embedding to each relation which needs parameters of $O(|R|d_r)$ where $|R|$ is the count of relations and $d$ is the embedding dimension, this network introduce only $O(d_r)$ parameters.

\xhdr{Relation-to-Entity Attention Network (R2E)} For relation-level aggregation, the entity representation aggregates from neighbor relations. In this way, relations are viewed as nodes which store the hidden information of entities. The hidden information may include the type constraint of head (or tail) entity, the max count of neighbor entities and so on, which are specific to the semantic of relation and are unable to capture via Entity-level Attention Network. So we propose Relation-to-Entity Attention Network to reconstruct entity representation. This process is named "Echo".

\begin{equation}
  \alpha_{ik}=\frac{\exp(\text{LeakyReLU}(\mathbf{a}^T[\mathbf{x}_i\| \mathbf{r}_k^h]))}{\sum_{\mathbf{r}_{k'}\in \mathcal{R}_{e_i}, }\exp(\text{LeakyReLU}(\mathbf{a}^T[\mathbf{x}_{i}\| \mathbf{r}_{k'}^h]))}
\end{equation}

\begin{equation}
  \mathbf{x}_i^{h,\mathbf{r}_k^h}=\sum_{\mathbf{r}_{k'}\in \mathcal{R}_{e_i} }\alpha_{ik} \mathbf{r}_k^h
\end{equation}

where $\mathcal{R}_{e_i}$ is the list, instead of the set, of relations related to head entity $e_i$ allowing duplicate relations specific to different tail entities, $\alpha_{ik}$ represents attention weight from the head part of relation $\mathbf{r}_k$ to head entity $e_i$, $\mathbf{a}$ is another $2d_r$-dimensional trainable vector, $d_r$ is the dimension of relation feature vector.

\subsubsection{Primitive Aggregation Network}
\label{PAN}

To obtain basic entity representations, we propose Primitive Aggregation Network (PAN) as follows:

\begin{equation}
  \mathbf{X}_e^{(\text{GCN})} = \text{FAN}(\text{GCN}(\mathbf{X}_e^{(\text{init})}), \mathbf{X}_e^{(\text{init})})
\end{equation}
\begin{equation}
  \begin{split}
    \mathbf{X}_e^{(\text{PAN})} & = \text{PAN}(\mathbf{X}_e^{(\text{init})}) \\
                                & = \text{FAN}(
                                      \text{EAN}(
                                        \text{FAN}(
                                          \text{EAN}(\mathbf{X}_e^{(\text{GCN})}),
                                          \mathbf{X}_e^{(\text{GCN})}
                                        )
                                      ),
                                      \mathbf{X}_e^{(\text{GCN})}
                                    )
  \end{split}
\end{equation}

In PAN, we don't take the type of relation into consideration because PAN focuses on basic entity representations related to simple, instead of complex, graph topological infomation. More fine topological infomation will be captured in next Echo Network. But we will show that PAN is more efficient than 2-layers GCN in ablation study.

\subsubsection{Echo Network}
\label{EN}

The Echo Network (EN) echos entity representations via two views of neighbor relation representations from primitive entity representations. Two views are the head view and the tail view of relation. They make different contributions to the semantic of relation because they are specific to head and tail entities respectively.

Given relation $\mathbf{r}_k$, we write the head view $\mathbf{r}_k^h$ and the tail view $\mathbf{r}_k^t$. Each view depends on related entities only. Similarly, the $i$-th entity $e_i$ representation can be projected to $\mathbf{x}_i^h=\mathbf{x}_i^{(\text{PAN})}\mathbf{W}^h$ and $\mathbf{x}_i^t=\mathbf{x}_i^{(\text{PAN})}\mathbf{W}^t$ where $\mathbf{W}^h,\mathbf{W}^t\in \mathbb{R}^{d_e\times d_r}$ are transformation matrices to transform from entity vector space to relation-specific vector space, $\mathbf{x}_i^{(\text{PAN})}$ is the entity embedding vector from PAN. We also keep $d_e = d_r$ to avoid destroying the shape.

Firstly, the entity information is sent to relation via Entity-to-Relation Attention Network. With $\mathbf{x}_i^h$ we get $\mathbf{r}_k^h$ and with $\mathbf{x}_i^t$ we get $\mathbf{r}_k^t$. Then the hidden information stored in relation is sent back to entity via Relation-to-Entity Attention Network. We can compute $\mathbf{x}_i^{h,\mathbf{r}_k^h}, \mathbf{x}_i^{t,\mathbf{r}_k^h}$ from $\mathbf{r}_k^h$ and $\mathbf{x}_i^{h,\mathbf{r}_k^t}, \mathbf{x}_i^{t,\mathbf{r}_k^t}$ from $\mathbf{r}_k^t$. Lastly, FAN is applied to automatically balance the information between $\mathbf{r}_k^h$ and $\mathbf{r}_k^t$. By concatenating, the echoed entity representation of $e_i$ is:

\begin{equation}
  \mathbf{x}_i^{(\text{EN})}=[\mathbf{x}_i^{(\text{PAN})} \| \text{FAN}(\mathbf{x}_i^{h,\mathbf{r}_k^h}, \mathbf{x}_i^{h,\mathbf{r}_k^t}) \| \text{FAN}(\mathbf{x}_i^{t,\mathbf{r}_k^h}, \mathbf{x}_i^{t,\mathbf{r}_k^t})]
\end{equation}

\subsubsection{Complete Aggregation Network}
\label{CAN}

With echoed entity embedding matrix $\mathbf{X}_e^{(\text{EN})}$ from EN, we design Complete Aggregation Network (CAN) to gather information from neighbors again.
To simplify things, another Entity-level Attention Network is applied, since EN pulls other levels (\eg relation-level) of information to entity level to construct the entity representation, while PAN ignores the relations acting on the entities.

\begin{equation}
  \mathbf{X}_e^{(\text{CAN})}=[\mathbf{X}_e^{(\text{EN})}\|\text{EAN}(\mathbf{X}_e^{(\text{EN})})]
\end{equation}

where $\mathbf{X}_e^{(\text{EN})} \in \mathbb{R}^{|E|\times (d_e+2d_r)} $ is the output of EN, $\mathbf{X}_e^{(\text{CAN})} \in \mathbb{R}^{|E|\times (2d_e+4d_r)} $ is the output of CAN. Note that $\mathbf{X}_e^{(\text{CAN})}$ is an alias of $\mathbf{X}_e^{(\text{Echo})}$.

\subsection{Align Loss}
\label{section:Loss}

The loss function is Hinge Loss with Manhattan distance $d(x,y)=\|x-y\|_1$:

\begin{equation}
  \label{eq:loss}
  \begin{split}
    L&=\sum_{(e_i,e_j)\in P^+,(e_i',e_j')\in P^-} \max (0, \lambda+d(\mathbf{x}_i,\mathbf{x}_j)-d(\mathbf{x}_i',\mathbf{x}_j'))) \\
    &+\sum_{(e_i',e_j')\in P_{\text{iter}}^-)} \max (0, \lambda-d(\mathbf{x}_i',\mathbf{x}_j'))
  \end{split}
\end{equation}

where $\lambda$ is a hyper-parameter of margin, $\mathbf{x}_i$ is entity $e_i$ embedding in $\mathbf{X}_e^{(\text{Echo})}$, $P^+$ is the set of positive samples, $P^-$ is the set of negative samples generated from $P^+$, $P_{\text{iter}}^-$ is the set of iterative negative samples.
These three sets are generated by our iterative strategy ABGS~\ref{section:ABGS}.

\subsection{Attribute-combined Bi-directional Global-filtered Strategy}
\label{section:ABGS}

We propose Attribute-combined Bi-directional Global-filtered Strategy (ABGS) to generate samples of high quality.

% \begin{algorithm}[t]
%   \caption{Attribute-combined Bi-directional Global-filtered Strategy.}
%   \label{alg:ABGS}
%   \KwIn{candidate entity sets $E_1',E_2'$, relation similarity matrix $S^{\text{rel}}\in \mathbb{R}^{|E_1'| \times |E_2'|}$.}
%   \KwOut{Iterative positive sample set $P_{\text{iter}}^+$ and iterative negative sample set $P_{\text{iter}}^-$.}
%   \BlankLine
%   Calculate attribute similarity $S^{\text{attr}}$ and value similarity $S^{\text{attr\_value}}$ following Section~\ref{section:ABGS:attr}\\
%   Calculate final similarity matrix following Formula~\ref{eq:attr-combine} \\
%   Calculate local alignment result as following step 4 to 7:\\
%   $P_{\text{local}}^1 = \{(e_1,\arg\max_{e_2}S(e_1,e_2))\;|\;e_1\in E_1', e_2\in E_2'\}$ \\
%   $P_{\text{local}}^2 = \{(\arg\max_{e_1}S(e_1,e_2),e_2)\;|\;e_1\in E_1', e_2\in E_2'\}$  \\
%   $P_{\text{local}}^+= P_{\text{local}}^1 \cap P_{\text{local}}^2$ \\
%   $P_{\text{local}}^- = P_{\text{local}}^1 \cup P_{\text{local}}^2 - P_{\text{local}}^+$ \\
%   Calculate global alignment result $P_{\text{global}}$ referring to~\cite{RAGA,CEA}\\
%   $P_{\text{iter}}^+ = P_{\text{local}}^+ \cap P_{\text{global}}$\\
%   $P_{\text{iter}}^- = P_{\text{local}}^- - P_{\text{global}}$ \\
%   \Return $P_{\text{iter}}^+,P_{\text{iter}}^-$
% \end{algorithm}

\subsubsection{Attribute Combination}

\paragraph{Relation-based similarity}
Relation-based similarity matrix $S^{\text{rel}}$ has $S^{\text{rel}}_{ij}=d(e_i,e_j)$ and $d(x,y)=\|x-y\|_1$ is distance function following Align Loss.

\paragraph{Attribute-based similarity}
\label{section:ABGS:attr}

Firstly we translate the names of the attributes to the same language (English) and then according to string matching measure(the Sorensen-Dice coefficient) as similarity, we filter aligned pairs of attributes with the top-1  similarity grater than a given threshold $\lambda$. With these comparable attributes, we denote $\text{Attr}(e_i)$ as the set of attributes of entity $e_i$. Lastly, we calculate the attribute-based similarity matrix $S^{\text{attr}}$ where $S_{ij}^{\text{attr}}=J(\text{Attr}(e_i), \text{Attr}(e_j))$, $e_i, e_j$are two entities from $\text{KG}_1$and $\text{KG}_2$ respectively, $J(A,B)=\frac{|A\cap B|}{|A\cup B|}$ represents the \emph{Jaccard} similarity of two sets $A$ and $B$.

\paragraph{Attribute-value-based similarity}

Firstly we get common attribute set $\text{Attr}_{ij}=\text{Attr}(e_i)\cap \text{Attr}(e_j)$. For each attribute $a \in \text{Attr}_{ij}$, the value similarity based on attribute $a$ of entity $e_i$ and $e_j$ is $S_{ij}^{(a)}=J(\text{Value}_a(e_i), \text{Value}_a(e_j))$ where $\text{Value}_a(e_i)$ is the value set of attribute $a$ of entity $e_i$. By averaging the value similarity of all attributes in $\text{Attr}_{ij}$, we get the attribute-value-based similarity matrix $S^{\text{attr\_value}}$ where $S^{\text{attr\_value}}_{ij}=\frac{1}{|\text{Attr}_{ij}|}\sum_{a\in \text{Attr}_{ij}}S_{ij}^{(a)}$.

\paragraph{Combination}

The final similarity matrix $S$ with hyper-parameters $\alpha_1,\alpha_2,\alpha_3 \in [0,1]$:

\begin{equation}
  \label{eq:attr-combine}
  S=\alpha_1 S^{\text{rel}}+\alpha_2 S^{\text{attr}}+\alpha_3 S^{\text{attr\_value}}
\end{equation}

\subsubsection{Bi-directional Global-filtered Strategy}

Firstly, according to final similarity matrix $S$ and nearest neighbor selection, we predict local alignment $P_{\text{left-to-right}}=\{(e_1, \arg\max_{e_2}S(e_1,e_2))\;|\;\text{Given}\;e_1\in E_1^{\text{test}}, \text{predict}\; e_2\in E_2^{\text{test}}\}$ (left-to-right prediction) and $P_{\text{right-to-left}}=\{(\arg\max_{e_1}S(e_1,e_2), e_2)\;|\;\text{Given}\;e_2\in E_2^{\text{test}}, \text{predict}\; e_1\in E_1^{\text{test}}\}$ (right-to-left prediction). Bi-directionally, local positive sample set is $P_{\text{local}}^+=P_{\text{left-to-right}} \cap P_{\text{right-to-left}}$ and local negative sample set is $P_{\text{local}}^-=P_{\text{left-to-right}} \cup P_{\text{right-to-left}} - P_{\text{local}}^+$.

Secondly, since $P_{\text{local}}^+$ and $P_{\text{local}}^-$ contains plenty of pairs that we are not able to determine if it is true or false, we filter it by one-to-one constraint. According to $S$ and nearest neighbor selection with one-to-one constraint, we predict global alignment $P_{\text{global}}$~\cite{RAGA, CEA}. Then we have the iterative positive samples $P_{\text{iter}}^+= P_{\text{local}}^+ \cap P_{\text{global}}$ and iterative negative samples $P_{\text{iter}}^-= P_{\text{local}}^- - P_{\text{global}}$.

Lastly, for next epoch training, $P_{\text{iter}}^+$ is viewed as true data and merged to train set, that is, the final positive sample set is $P^+=P_{\text{seed}} \cup P_{\text{iter}}^+$. We adopt the nearest neighbour sampling to sample $k$ negative samples as $P^-$ from $P^+$ according to $S$.

\section{Experiments}
\subsection{Experimental Settings}

\paragraph{Dataset and Metrics}
\label{sec:exp:dataset}
Three cross-lingual datasets: ZH-EN(Chinese to English), JA-EN(Japanese to English), and FR-EN(French to English) from DBP15K~\cite{JAPE} were employed in this experiment, shown in Table~\ref{dataset}. we randomly split 30\% of pre-aligned pairs for training and keep 70\% for testing. The reported performance is the average of five independent runs and the train/test datasets are shuffled before training. We use Hits@k and Mean Reciprocal Rank(MRR) following previous work~\cite{JAPE}. For Global Alignment, there is only \emph{Hits@1} since it makes one-to-one alignment. For all metrics, the higher, the better.

\begin{table*}
  \caption{Statistical data of real-world datasets.}\label{dataset}
  \centering
  \begin{tabular}{cccccccc}
    \toprule[1pt]
    \#Dataset              & \#Lang & \#Ent   & \#Rel & \# Att & \#Rel Triples & \#Attr Triples & \#Ent Seeds             \\
    \midrule[1pt]
    \multirow{2}{*}{ZH-EN} & ZH     & 66,469  & 2,830 & 8,113  & 153,929       & 379,684        & \multirow{2}{*}{15,000} \\
                           & EN     & 98,125  & 2,317 & 7,173  & 237,674       & 237,674        &                         \\
    \midrule[1pt]
    \multirow{2}{*}{JA-EN} & JA     & 65,744  & 2,043 & 5,882  & 164,373       & 354,619        & \multirow{2}{*}{15,000} \\
                           & EN     & 95,680  & 2,096 & 6,066  & 233,319       & 497,230        &                         \\
    \midrule[1pt]
    \multirow{2}{*}{FR-EN} & FR     & 66,858  & 1,379 & 4,547  & 192,191       & 528,665        & \multirow{2}{*}{15,000} \\
                           & EN     & 105,889 & 2,209 & 6,422  & 278,590       & 576,543        &                         \\
    \bottomrule[1pt]
  \end{tabular}
\end{table*}

\paragraph{Implementation Details}
\label{sec:exp:impl}

For fair comparison, we use the initial entity embeddings from RDGCN~\cite{RDGCN}, which translates all entity names to English via Google Translate and then uses Glove~\cite{Glove} to construct the initial entity embeddings (dimension $d_e=300$).
Following~\cite{RAGA,RREA}, we choose learning rate $r=0.001$, margin $\lambda=3$, and the negative sample number $k=5$.
Other hyper-parameters are chosen from the following candidate sets: dropout rate $\mu=0.05\in\{0.05, 0.1, 0.15, 0.2, 0.25\}$, attribute combination weight $(\alpha_1,\alpha_2,\alpha_3) =(0.1, 0.5, 0.4)\in \{(0.33, 0.33, 0.33), (0.1, 0.45, 0.45), (0.5, 0.25, 0.25), (0.1, 0.5, 0.4)\}$ (the candidate set is inspired by~\cite{EMGCN}), the number of epochs for bootstrapping and updating negative samples $p=10 \in \{5,10,15,20,25\}$.
We used two GTX1080 graphic cards. We implemented our model with \emph{PyTorch} and used Adam as a gradient optimizer. Source code is available at \url{https://github.com/LinXueyuanStdio/EchoEA}.

\paragraph{Baselines}

We focus on GNN-based methods and semi-supervised methods. Therefore, we compare to the following methods which have been introduced in Section~\ref{sec:relatedwork}:

\begin{itemize}
  \item \emph{Basic}: This kind of methods only uses relation triples. GCN-Align~\cite{GCN-Align}, RDGCN~\cite{RDGCN}, MRAEA~\cite{MRAEA}, RREA(b) (basic version of RREA~\cite{RREA}), RAGA(b) (basic version of RAGA~\cite{RAGA}).
  \item \emph{Semi-supervised}: This kind of methods introduces semi-supervision to generate extra data. BootEA~\cite{BootEA}, MRAEA~\cite{MRAEA}, RREA(s) (semi-supervised version of RREA~\cite{RREA}).
  \item \emph{Complete}: This kind of methods use their best experimental settings. RREA~\cite{RREA}. EMGCN~\cite{EMGCN}. RAGA~\cite{RAGA}.
\end{itemize}

\paragraph{Model Variants}

To compare with other methods and determine how important each design choice of our model is, we provide variants of \emph{EchoEA} as follow:

\begin{itemize}
  \item \textbf{EchoEA(b)}: Only relation triples are used with Local Alignment.
  \item \textbf{EchoEA(g)}: Only relation triples are used with Global Alignment.
  \item \textbf{EchoEA(s)}: EchoEA(b) with semi-supervised strategy ABGS~\ref{section:ABGS}.
  \item \textbf{EchoEA}: The complete EchoEA, which is EchoEA(s) with Global Alignment.
  \item \textbf{w/o PAN}: EchoEA(b) without Primitive Aggregation Network~\ref{PAN}.
  \item \textbf{w/o EN}: EchoEA(b) without Echo Network~\ref{EN}.
  \item \textbf{w/o CAN}: EchoEA(b) without Complete Aggregation Network~\ref{CAN}.
\end{itemize}

\subsection{Experimental Results and Analysis}

\begin{table*}
  \caption{Overall performance of entity alignment.}
  \label{result}
  \centering
  \begin{threeparttable}
    \begin{tabular}{llllllllll}
      \toprule[1pt]
                         & \multicolumn{3}{c}{ZH-EN} & \multicolumn{3}{c}{JA-EN} & \multicolumn{3}{c}{FR-EN}                                                                                                              \\
      Methods\tnote{1}   & H@1                       & H@10                      & MRR                       & H@1                & H@10             & MRR            & H@1             & H@10           & MRR            \\
      \midrule[1pt]
      GCN-Align          & 41.3                      & 74.4                      & 0.549                     & 39.9               & 74.5             & 0.546          & 37.3            & 74.5           & 0.532          \\
      RDGCN              & 70.8                      & 84.6                      & -                         & 76.7               & 89.5             & -              & 88.6            & 95.7           & -              \\
      MRAEA              & 63.5                      & 88.2                      & 0.729                     & 63.6               & 88.7             & 0.731          & 66.6            & 91.2           & 0.764          \\
      RREA(b)            & 71.5                      & 92.9                      & 0.794                     & 71.3               & 93.3             & 0.793          & 73.9            & 94.6           & 0.816          \\
      EMGCN(b)           & 72.5                      & 87.5                      & 0.778                     & 78.0               & 92.0             & 0.831          & 89.2            & 97.9           & 0.925          \\
      RAGA(b)            & 79.8                      & 93.0                      & 0.847                     & 83.1               & 95.0             & 0.875          & 91.4            & 98.3           & 0.940          \\
      \hdashline
      Init-Emb\tnote{2}  & 57.5                      & 68.9                      & 0.615                     & 65.0               & 75.4             & 0.688          & 81.8            & 88.8           & 0.843          \\
      w/o PAN            & 74.76                     & 86.63                     & 0.791                     & 79.75              & 90.21            & 0.835          & 91.03           & 96.53          & 0.931          \\
      w/o EN             & 80.11                     & 93.34                     & 0.848                     & 84.83              & 95.42            & 0.887          & 93.06           & 98.58          & 0.952          \\
      w/o CAN            & 77.42                     & 89.96                     & 0.819                     & 82.53              & 93.13            & 0.864          & 92.96           & 98.12          & 0.950          \\
      \hdashline
      \textbf{EchoEA(b)}          & \textbf{82.30}            & \textbf{93.93}            & \textbf{0.865}            & \textbf{86.09}     & \textbf{95.70}   & \textbf{0.897} & \textbf{93.92}  & \textbf{98.90} & \textbf{0.958} \\
      \textbf{EchoEA(g)}          & \textbf{89.13}            & -                         & -                         & \textbf{93.22    } & -                & -              & \textbf{97.67 } & -              & -              \\
      \midrule[1pt]
      BootEA             & 62.9                      & 84.8                      & 0.703                     & 62.2               & 85.4             & 0.701          & 65.3            & 87.4           & 0.731          \\
      MRAEA              & 75.7                      & 93.0                      & 0.827                     & 75.8               & 93.4             & 0.826          & 78.1            & 94.8           & 0.849          \\
      RREA(s)            & 80.1                      & 94.8                      & 0.857                     & 80.2               & 95.2             & 0.858          & 92.7            & 96.6           & 0.881          \\
      \hdashline
      Attr-Sim\tnote{3}  & 13.34                     & 13.81                     & 0.190                     & 5.30               & 11.23            & 0.089          & 0.79            & 5.08           & 0.025          \\
      Value-Sim\tnote{3} & 41.95                     & 46.55                     & 0.498                     & 34.16              & 54.78            & 0.414          & 63.93           & 79.47          & 0.697          \\
      \hdashline
      \textbf{EchoEA(s)}          & \textbf{93.03}            & \textbf{97.21}            & \textbf{0.945}            & \textbf{94.20	}     & \textbf{ 98.13 } & \textbf{0.957} & \textbf{97.91}  & \textbf{99.53} & \textbf{0.987} \\
      \midrule[1pt]
      RREA               & 82.2                      & -                         & -                         & 91.8               & -                & -              & 96.3            & -              & -              \\
      EMGCN              & 86.25                     & -                         & -                         & 86.63              & -                & -              & 93.95           & -              & -              \\
      RAGA               & 87.3                      & -                         & -                         & 90.9               & -                & -              & 96.6            & -              & -              \\
      \textbf{EchoEA}             & \textbf{94.99}            & -                         & -                         & \textbf{96.46}     & -                & -              & \textbf{98.93}  & -              & -              \\
      \bottomrule[1pt]
    \end{tabular}
    \begin{tablenotes}
      \footnotesize
      \item[1] All results of compared methods are taken from their original papers since their variants share the same settings. For better comparison, the results are grouped by \emph{Basic} (with ablation results of EchoEA), \emph{Semi-supervised} and \emph{Complete}.
      \item[2] Init-Emb: The initial entity embedding widely applied in RDGCN, RREA, EMGCN and RAGA.
      \item[3] Attr-Sim, Value-Sim: they represent the results generated from similarity matrix based on attribute and attribute value respectively, which is taken from EMGCN.
    \end{tablenotes}
  \end{threeparttable}
\end{table*}

\paragraph{EchoEA vs. Basic}

For basic methods, GCN-Align performs worst owing to ignoring the relation types and count. RDGCN and MRAEA are better because of their further utilization of relation information for entities. EMGCN filters the noise in propagation with multi-order GCN layers. In order to leverage more relation information, RREA explicitly models the interaction between entities and relations, thus getting better than methods before. Symmetrically, RAGA takes the contribution of entities to relations into consideration and implicitly models the interaction between entities and relations, which makes contributions to higher performance.
However, they still suffers from limitation to aggregation layers and oversmoothing.
To the opposite, our Echo model is built with four levels attention networks and it further utilises relation information by echoing information between entities and relations.
Therefore, EchoEA(b) performs the best of all.

\paragraph{EchoEA vs. Semi-supervision}

Compared with those methods which introduces semi-supervision, EchoEA(s) also performs the best due to the brilliant core encoder network \emph{Echo} and enhanced iterative strategy \emph{ABGS}.
Compared to EchoEA(b), EchoEA(s) generates extra training data via semi-supervision which brings 3.99-11.27\% improvement on \emph{Hits@1}.

\paragraph{EchoEA vs. All}

Considering the best results of all baselines, obviously, our model consistently outperforms all competing basic methods on all the evaluation metrics.
The highest \emph{Hits@1} reaches 97.91\% in FR\_EN. The average \emph{Hits@1} is approximate 96\% on these three data sets, significantly showing that our framework outperforms the state-of-the-art methods.

\paragraph{Effect of Each Component}

Taking w/o PAN, w/o EN and w/o CAN into account, we can see that PAN, EN and CAN all improve the performance significantly. The order of importance of each component is PAN>CAN>EN. In addition, it should be highlighted that \emph{w/o EN} has already outperformed RAGA~\cite{RAGA}. The result shows the efficience of each component of Echo model.

\paragraph{Quality of generated data from ABGS}

To measure the quality of iterative samples, we use entity utilization rate $r_u$, false positive rate $r_{p}$ and false negative rate $r_n$.

\begin{equation}
  r_u = \frac{|P_{\text{iter}}^+|+|P_{\text{iter}}^-|}{|P_{\text{test}}|}
\end{equation}
\begin{equation}
  r_p = \frac{|P_{\text{iter}}^+ - P_{\text{test}}|}{|P_{\text{iter}}^+|}
\end{equation}
\begin{equation}
  r_n = \frac{|P_{\text{iter}}^- \cap P_{\text{test}}|}{ |P_{\text{iter}}^- |}
\end{equation}

\begin{figure*}
  \label{sec:appendix:utilization}
  \centering
  \includegraphics[width=\textwidth]{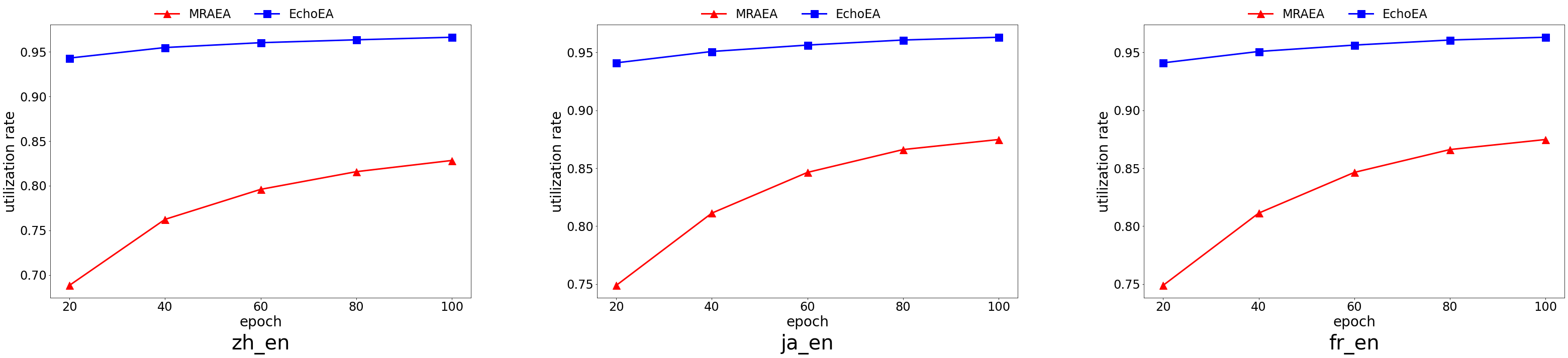}
  \caption{Entity utilization rate of EchoEA and MRAEA. The higher, the better.}
\end{figure*}

\begin{figure*}
  \label{sec:appendix:false_rate}
  \centering
  \includegraphics[width=\textwidth]{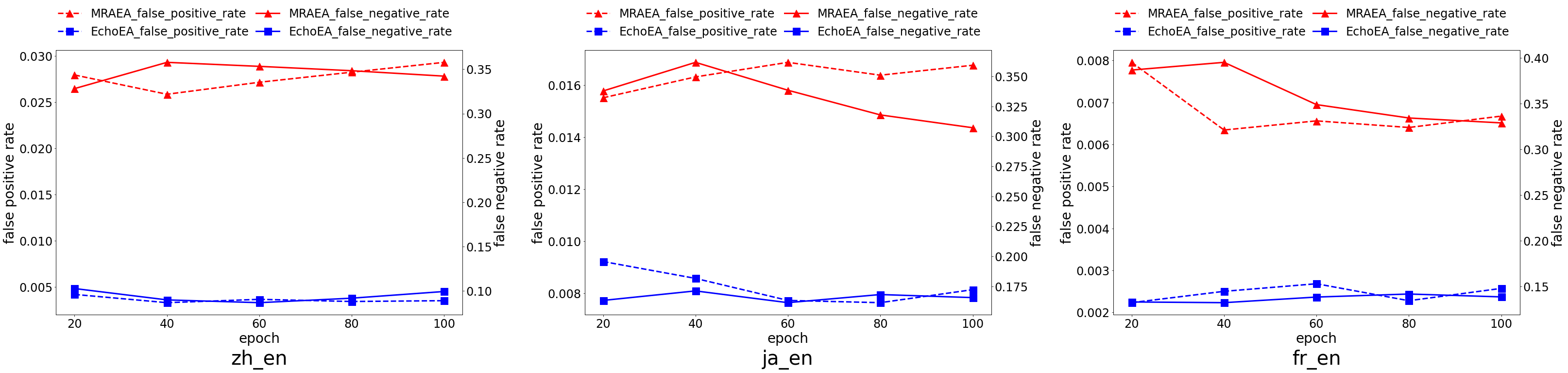}
  \caption{False positive rate and false negative rate of EchoEA and MRAEA. The lower, the better.}
\end{figure*}

Here $P_{\text{test}}$ represents the true alignment. $P_{\text{iter}}^+$ and $P_{\text{iter}}^-$ represent the iterative positive samples set and iterative negative samples set respectively. We take the first 100 training epochs of EchoEA and MRAEA~\cite{MRAEA} to analysis. Results are shown in Figure~\ref{sec:appendix:utilization} and Figure~\ref{sec:appendix:false_rate}.


From Figure~\ref{sec:appendix:utilization}, we can see that EchoEA and MRAEA all have an upward trend of entity utilization rate during the first 100 training epochs.
However, the entity utilization rate of EchoEA is around 95\% at three dataset.
It is obviously higher than MRAEA, which shows that EchoEA generates more pairs.

From Figure~\ref{sec:appendix:false_rate}, the false positive rate and the false negative rate of EchoEA and MRAEA both remain steady. Compared with MRAEA, EchoEA brings about 50\% decrease in false positive rate and false negative rate.

The above analysis has brought us to the conclusion that ABGS can provide more iterative samples with higher quality.

\section{Conclusion}\label{sec:conclusion}

In this paper, we propose a novel EA framework \emph{EchoEA} for cross-lingual entity alignment. To model the interaction between entities and relations, we propose \emph{Echo} as entity encoder to digest information in relation triples. The Echo is 5-layers and it is deeper than previous methods. In addition, we propose Attribute-combined Bi-directional Global-filtered Strategy (\emph{ABGS}) to extend datasets. Our method outperforms the state-of-the-art GNN-based entity alignment methods by a large margin across three real-world cross-lingual datasets.

% \paragraph{Limitation}
% \label{Limitation}

% On the one hand, our method depends on machine translation technology to translate entity names, attribute names and attribute value names to the same language. In our framework, machine translation is not integrated to end-to-end training. So language-model-based methods may outperform our approach due to larger parameters. But instead, we are confident to outperform GNN-based SOTA methods.
% On the other hand, the false pairs generated by ABGS are permanently harmful during training if the GNN network remenber the false data.

% \input{paper_section/061_broader}

% or


% TODO 先去掉附录
% \appendices

% \input{paper_section/070_appendix}

\section*{Acknowledgment}
This work was supported in part by the National Science Foundation of China (Grant No.61902034); Engineering Research Center of Information Networks, Ministry of Education of China

\ifCLASSOPTIONcaptionsoff
  \newpage
\fi



\bibliographystyle{IEEEtranN}
\bibliography{paper_references/bibliography}


% \end{thebibliography}

\begin{IEEEbiography}[
  {\includegraphics[width=1in,height=1.25in,clip,keepaspectratio]{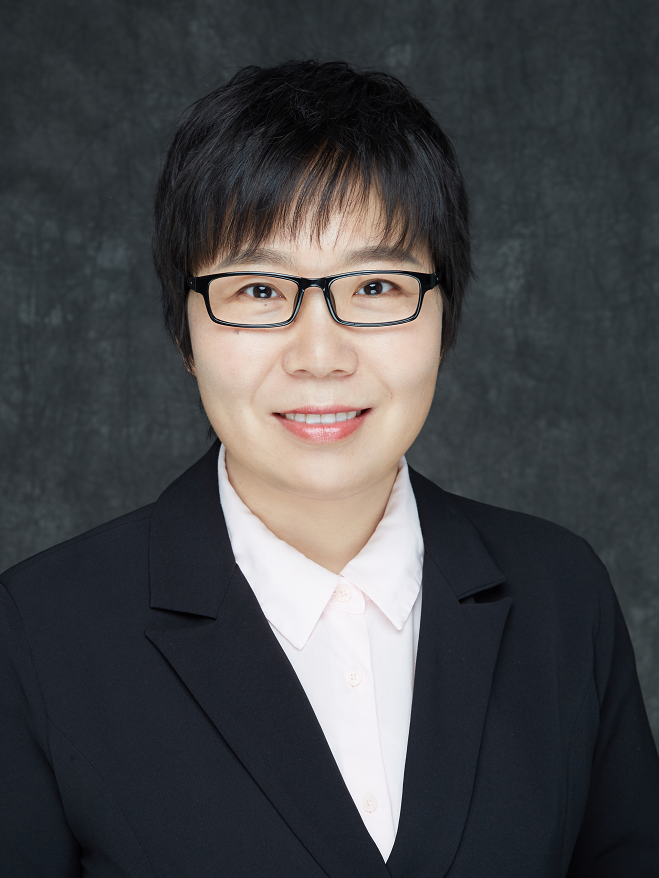}}
]{Haihong E}
(1982- ), female (Han), native of Liaoning Province, China. associate professor and Ph.D., School of Computer Science, Beijing University of Posts and Telecommunications, major research areas include deep learning, knowledge graph, natural language processing, big data and artificial intelligence.
\end{IEEEbiography}

\begin{IEEEbiographynophoto}{Xueyuan Lin}
(1998- ), male (Han), native of Guangdong Province, China. Beijing University of Posts and Telecommunications, School of Computer Science, postgraduate student, the main research areas include deep learning, knowledge graph, natural language processing, big data and artificial intelligence.
\end{IEEEbiographynophoto}

\begin{IEEEbiographynophoto}{Wenyu Song}
(1998- ), male (Han), native of Shandong Province, China. Beijing University of Posts and Telecommunications, School of Computer Science, postgraduate student, the main research areas include deep learning, graph neural network, knowledge graph.
\end{IEEEbiographynophoto}

\begin{IEEEbiography}[
  {\includegraphics[width=1in,height=1.25in,clip,keepaspectratio]{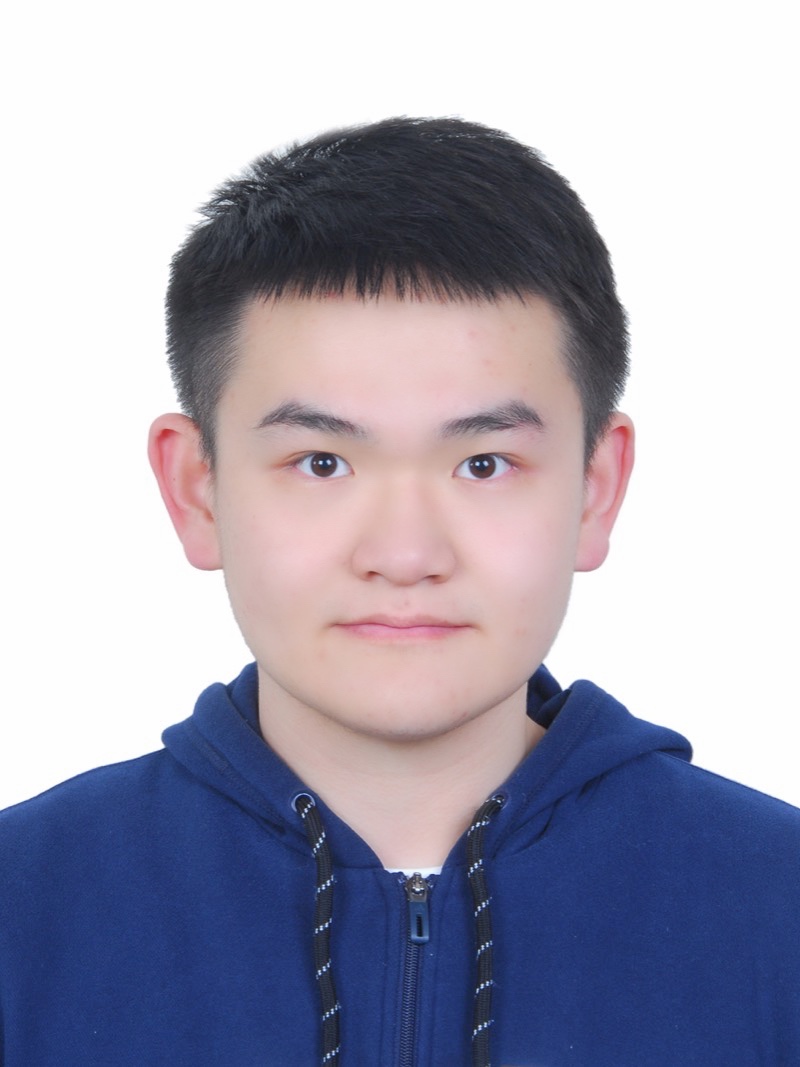}}
]{Haoran Luo}
(1998- ), male (Han), native of Liaoning Province, China. Beijing University of Posts and Telecommunications, School of Computer Science, postgraduate student, the main research areas include machine learning, deep learning, natural language processing, knowledge graph, computer vision and Big data.
\end{IEEEbiography}

% \end{IEEEbiography}


% % insert where needed to balance the two columns on the last page with
% %\newpage



%\vfill


\end{document}